\documentclass[runningheads]{llncs}
\usepackage[T1]{fontenc}
\usepackage{graphicx}
\usepackage{amsmath}
\usepackage{amssymb}
\usepackage{cite}
\usepackage{booktabs}
\usepackage{multirow}
\usepackage[ruled,vlined]{algorithm2e}
\usepackage{tikz}
\usetikzlibrary{positioning, arrows.meta, backgrounds}

\usepackage{xcolor}

\begin{document}

\title{Physics-Informed Multi-Agent Coordination for Hospital Patient Flow Optimization}
\titlerunning{Physics-Informed Multi-Agent Coordination in Hospital Networks}

\author{Guoqing Zhang\orcidID{0009-0007-6956-0814} \and
    Rafik Hadfi\orcidID{0000-0003-2352-1936} \and
    Takayuki Ito\orcidID{0000-0001-5093-3886}}
\authorrunning{G. Zhang et al.}
\institute{Graduate School of Informatics, Kyoto University, Kyoto, Japan \\
    \email{zhang.guoqing.27t@st.kyoto-u.ac.jp, \{rafik.hadfi, ito\}@i.kyoto-u.ac.jp}}

\maketitle

\vspace{-1em}
\begin{abstract}
 Efficient coordination of patient flow between autonomous hospital departments is critical to mitigate overcrowding and balance resource utilization. While queueing theory, specifically open Baskett-Chandy-Muntz-Palacios (BCMP) networks, provides an interpretable topology for healthcare operations, analytical models rely on stationary assumptions and fixed routing matrices that degrade under state-dependent real-world dynamics. In contrast, centralized reinforcement learning approaches struggle to accommodate the decentralized structure of hospital governance, where individual clinical departments function with localized observations, heterogeneous resources, and divergent operational objectives. In this paper, we present a Multi-Agent Systems (MAS) framework titled \emph{Physics-Informed Multi-Agent Coordination}, which embeds empirically calibrated BCMP queueing topologies as physical priors within a decentralized multi-agent reinforcement learning architecture. Formulated as a Decentralized Partially Observable Markov Decision Process (Dec-POMDP) under coupled resource constraints, our method enables autonomous departmental agents to cooperatively negotiate patient routing and dynamic service scaling. To mitigate environmental non-stationarity without inducing excessive communication overhead, agents exchange localized action fingerprints along network edges and optimize a spatially decomposed reward structure. Empirical evaluations driven by real-world MIMIC-IV patient trajectories indicate that this cooperative multi-agent approach substantially reduces cumulative system delay compared to static Markovian approximations, heuristic dispatching, and independent multi-agent baselines, while maintaining clinical safety constraints.

\end{abstract}

\vspace{-1em}
\section{Introduction}

Modern large-scale hospitals are intrinsically decentralized organizations characterized by simultaneous interactions among autonomous clinical units, such as the Emergency Department (ED), Intensive Care Units (ICUs), surgical suites, and general inpatient wards. These specialized departments continuously manage heterogeneous patient populations while competing for finite, highly shared medical resources including physical beds, clinical staff, and diagnostic equipment. As global emergency healthcare demand continuously surges, hospitals face acute pressures to safely and expediently coordinate patient movement across clinical stages \cite{armony2015patient,vainieri2020waiting}. When inter-departmental coordination breaks down, localized congestion cascades rapidly into emergency boarding, protracted waiting times, and systemic operational failure.

Optimizing hospital-wide patient flow presents a challenging multi-agent coordination problem. Existing healthcare operations research has traditionally approached this problem through centralized heuristic scheduling, mixed-integer programming, static capacity planning, or analytical queueing theory \cite{long2018boarding,laskowski2009models,lei2014mixed,barz2015elective,schmidt2013decision}. In particular, considerable effort has focused on decision support and matheuristics for automated patient-to-bed assignment under stochastic stay lengths \cite{bekker2011scheduling,thomas2013automated,vancroonenburg2016study,guido2018efficient}. Among analytical models, open Baskett-Chandy-Muntz-Palacios (BCMP) queueing networks are frequently employed to represent multi-class patient progression, probabilistic transitions, and diverse service disciplines under steady-state conditions. However, analytical BCMP formulas assume stationary arrival rates, fixed transition probabilities, and independent node queues. In clinical practice, departments operate with finite bed capacities; when a downstream unit reaches saturation, transfers from upstream wards are blocked. Under these state-dependent routing dynamics, the mathematical assumptions required for product-form equilibrium fail, limiting the applicability of static queueing solutions for dynamic operational intervention.

From a Multi-Agent Systems (MAS) perspective, a centralized control architecture (whether driven by global mathematical heuristics or monolithic deep reinforcement learning, such as PPO or DQN) is structurally ill-suited for hospital-wide clinical scheduling. In operational reality, clinical departments operate as functionally and administratively decentralized entities governed by supervisory hospital management boards \cite{funhiro2020standardization}, characterized by local observations, local operational resources, and naturally conflicting clinical objectives. For example, the Emergency Department strives to accelerate patient discharge to prevent reception overflow; the ICU operates under strict admission thresholds to reserve high-acuity ventilators for critical surgeries; and general wards aim to maintain stable bed occupancy without exhausting nursing shifts. Attempting to force these distinct domains under a centralized control algorithm creates computational bottlenecks, communication overhead, and organizational friction. Instead, patient flow coordination must be modeled as decentralized decision-making under coupled resource constraints, which represents a canonical paradigm in advanced MAS research.

The broader availability of longitudinal Electronic Health Record (EHR) datasets, such as MIMIC-IV \cite{johnson2023mimic}, facilitates empirical, data-driven system modeling. Rather than discarding queueing mechanics in favor of unconstrained black-box simulators, we utilize BCMP networks to establish topological structure and traffic conservation bounds. In this paper, we formulate a Physics-Informed Multi-Agent Coordination framework. Within this architecture, the empirical BCMP network acts as a structural physical prior that defines network connectedness, clinical transition pathways, and server capacity thresholds, while cooperative multi-agent reinforcement learning (MARL) negotiates patient routing and capacity scaling across localized clinical nodes.

The core technical and empirical contributions of this work are summarized as follows:
\begin{itemize}
    \item \textbf{Physics-Informed MAS Framework:} We integrate queueing theory with multi-agent control by leveraging empirical BCMP network dynamics calibrated on MIMIC-IV as structural physical priors. This formulation provides topology and resource capacity boundaries for multi-agent coordination in finite-server environments.
    \item \textbf{Topology-Aware Decentralized Coordination Mechanism:} We formulate hospital resource management as a Dec-POMDP and implement an action-fingerprint communication protocol. By sharing smoothed policy representations along topological edges, adjacent agents mitigate partial observability and environmental non-stationarity without global communication overhead.
    \item \textbf{Safety-Constrained Cooperative Routing Policy:} We formulate patient flow diversion as an emergent multi-agent routing policy driven by localized negotiations between adjacent departments. To maintain clinical protocol integrity and prevent hazardous patient routing, policy action spaces are modulated by an online expert action masking mechanism derived from historical clinician transition trajectories.
    \item \textbf{Empirical Performance Analysis:} Through clinical dataset calibrations and ablation experiments, we show that cooperative multi-agent action decomposition helps suppress non-linear congestion cascades. Our approach achieves an order-of-magnitude reduction in cumulative patient delay compared to static analytical approximations, while consistently outperforming heuristic dispatchers and independent MARL baselines.
\end{itemize}

\vspace{-1em}
\section{Related Work}

\subsection{BCMP Queueing Networks in Healthcare Operations}
The application of BCMP queueing networks to hospital patient flow represents a mature lineage within medical operations research. Classical works, such as Armony et al. \cite{armony2015patient}, provided analytical foundations for characterizing multi-stage inpatient trajectories and establishing theoretical occupancy thresholds. Extending this direction, Jebbor et al. \cite{jebbor2019approach} utilized empirical hospital statistics to jointly optimize human staffing levels and physical bed configurations within an open queueing framework, demonstrating significant waiting time reductions upon real-world deployment. Recent efforts by Mizuno et al. \cite{mizuno2026resource} advanced resource allocation planning by incorporating heterogeneous service times into steady-state analytical bounds.

While analytical BCMP hospital models offer structural clarity, they assume stationary equilibrium behavior. To maintain a tractable product-form distribution, these models require that patient transition probabilities between departments remain fixed and independent of system load. In realistic clinical settings, finite departmental server capacity induces physical blocking: when downstream diagnostic or surgical units saturate, transfer pathways close and upstream queues accumulate non-linearly. Under such state-dependent transition dynamics, Markovian product-form independence holds no longer, making stationary mathematical derivations unsuitable for real-time control intervention.

\subsection{Multi-Agent Systems and Collaborative Control}
To manage interconnected dynamic environments without relying on intractable analytical solvers, Multi-Agent Systems (MAS) have emerged as a dominant paradigm in intelligent decision-making. In domains such as traffic network signal control, logistics dispatching, and power grids, formulating resource allocation as a Decentralized Partially Observable Markov Decision Process (Dec-POMDP) allows independent actors to make localized decisions under global systemic objectives \cite{chu2019multi}. In recent years, multi-agent reinforcement learning (MARL) has gained momentum across decentralized healthcare applications, demonstrating potential in collaborative clinical decision-making \cite{ekpo2025skill}, medical resource allocation under imperfect information \cite{hao2022hierarchical}, UAV-enabled Internet of Medical Things (IoMT) infrastructures \cite{seid2023multiagent,sheikh2025advancing}, and emergency transportation routing \cite{lv2026multi}. However, these existing healthcare MARL architectures predominantly treat the clinical ecosystem as an abstract black-box simulator. By ignoring the fundamental queueing mechanics and physical capacity limits inherent to hospital dynamics, independent learning actors fail in tightly coupled wards due to environmental non-stationarity; when multiple departmental agents simultaneously update their routing preferences without communication, localized optimization traps and oscillating downstream bottlenecks emerge.

Recent advances in MARL emphasize structured collaborative mechanisms, such as Value-Decomposition Networks (VDN), QMIX, and Multi-Agent Actor-Critic architectures equipped with spatial communicative layers or action fingerprints \cite{chu2019multi}. While these collaborative algorithms excel in unconstrained simulation benchmarks, deploying multi-agent control directly to safety-critical healthcare networks remains largely unaddressed. Unlike games or abstract routing grids, hospital dispatching cannot tolerate unconstrained structural exploration or myopic global optimization that sacrifices individual patient pathways. Our study bridges this disconnect by demonstrating how empirical BCMP queueing topologies can function as structural domain physics, guiding multi-agent cooperation, bounding negotiation spaces, and ensuring verifiable clinical safety.

\vspace{-1em}
\section{Physics Prior: BCMP Queueing Topology and Constraints}

Rather than utilizing queueing derivations to calculate steady-state operational limits, we repurpose the mathematical architecture of open BCMP networks as the foundational \textit{physics engine} and structural boundary condition for multi-agent interaction.

\subsection{Network Topology and Traffic Conservation}

We model the physical infrastructure of a hospital as a directed topological network consisting of $M$ discrete clinical nodes $\mathcal{N} = \{1, \dots, i, \dots, M\}$, representing departments such as the ED, Surgery, Intensive Care Units, and General Medicine wards. Arriving patients are stratified into clinical job classes $\mathcal{K} = \{1, \dots, k, \dots, K\}$.

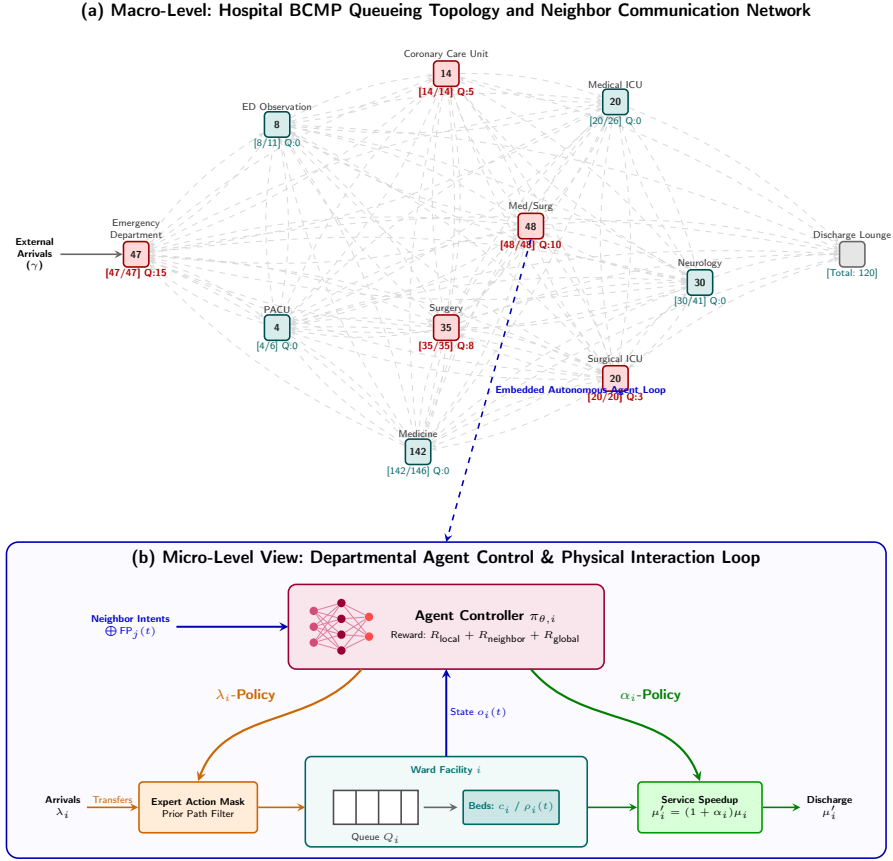
\begin{figure}[htbp]
    \centering
    \vspace{-0.5em}
    \resizebox{0.90\textwidth}{!}{
        \begin{tikzpicture}[
                font=\sffamily\scriptsize,
                >=stealth,
                base_node/.style={rectangle, rounded corners=2pt, thick, minimum size=5.0mm, inner sep=1.2pt, font=\sffamily\scriptsize\bfseries},
                congested_node/.style={base_node, fill=red!15, draw=red!60!black, text=black!90},
                normal_node/.style={base_node, fill=teal!15, draw=teal!60!black, text=black!90},
                discharge_node/.style={base_node, fill=black!10, draw=black!60, text=black!90},
                dense_edge/.style={->, draw=black!15, dashed, very thin},
                main_edge/.style={->, draw=black!60, thick},
                lbl_above/.style={align=center, text=black!85, font=\sffamily\scriptsize\bfseries, yshift=-0.5mm},
                lbl_below/.style={align=center, text=teal!85!black, font=\sffamily\scriptsize, yshift=0.5mm},
                lbl_alert/.style={align=center, text=red!75!black, font=\sffamily\scriptsize\bfseries, yshift=0.5mm}
            ]
            \node[font=\sffamily\small\bfseries, text=black] at (0, 3.6) {(a) Macro-Level: Hospital BCMP Queueing Topology and Neighbor Communication Network};

            \node[congested_node, label={[lbl_above]above:Emergency\\Department}, label={[lbl_alert]below:[47/47] Q:15}] (ED) at (-5.5, 0) {47};
            \node[normal_node, label={[lbl_above]above:ED Observation}, label={[lbl_below]below:[8/11] Q:0}] (EDObs) at (-3, 2.1) {8};
            \node[normal_node, label={[lbl_above]above:PACU}, label={[lbl_below]below:[4/6] Q:0}] (PACU) at (-3, -1.1) {4};
            \node[congested_node, label={[lbl_above]above:Coronary Care Unit}, label={[lbl_alert]below:[14/14] Q:5}] (CCU) at (0, 2.8) {14};
            \node[congested_node, label={[lbl_above]above:Surgery}, label={[lbl_alert]below:[35/35] Q:8}] (Surgery) at (0, -1.1) {35};
            \node[normal_node, label={[lbl_above]above:Medicine}, label={[lbl_below]below:[142/146] Q:0}] (Medicine) at (-0.5, -2.7) {142};
            \node[congested_node, label={[lbl_above]above:Med/Surg}, label={[lbl_alert]below:[48/48] Q:10}] (MedSurg) at (1.5, 0.5) {48};
            \node[normal_node, label={[lbl_above]above:Medical ICU}, label={[lbl_below]below:[20/26] Q:0}] (MICU) at (3, 2.3) {20};
            \node[congested_node, label={[lbl_above]above:Surgical ICU}, label={[lbl_alert]below:[20/20] Q:3}] (SICU) at (3, -1.7) {20};
            \node[normal_node, label={[lbl_above]above:Neurology}, label={[lbl_below]below:[30/41] Q:0}] (Neurology) at (4.5, -0.5) {30};
            \node[discharge_node, label={[lbl_above]above:Discharge Lounge}, label={[lbl_below]below:[Total: 120]}] (Discharge) at (7.2, 0) {};
            \node[align=center, font=\sffamily\scriptsize\bfseries] (Arrival) at (-7.3, 0) {External\\Arrivals\\($\gamma$)};

            \begin{scope}[on background layer]
                \foreach \i in {ED, EDObs, CCU, MICU, PACU, Medicine, SICU, MedSurg, Surgery, Neurology} {
                        \foreach \j in {ED, EDObs, CCU, MICU, PACU, Medicine, SICU, MedSurg, Surgery, Neurology, Discharge} {
                                \edef\tempi{\i}\edef\tempj{\j}
                                \ifx\tempi\tempj\else
                                    \draw[dense_edge] (\i) to[bend left=10] (\j);
                                \fi
                            }
                    }
            \end{scope}
            \draw[main_edge] (Arrival) -- (ED);

            \node[draw=blue!70!black, thick, rounded corners=6pt, fill=blue!2, minimum width=15.6cm, minimum height=4.2cm] (AgentPanel) at (0, -6.1) {};
            \node[font=\sffamily\small\bfseries, text=black] at (0, -4.2) {(b) Micro-Level View: Departmental Agent Control \& Physical Interaction Loop};

            \node[rectangle, rounded corners=4pt, thick, draw=purple!80!black, fill=purple!10, minimum width=5.8cm, minimum height=1.35cm] (Brain) at (0, -5.1) {};

            \begin{scope}[shift={(-2.0, -5.1)}, scale=0.32]
                \node[circle, fill=purple!70, inner sep=1.8pt] (I1) at (-1, -0.8) {};
                \node[circle, fill=purple!70, inner sep=1.8pt] (I2) at (-1, 0) {};
                \node[circle, fill=purple!70, inner sep=1.8pt] (I3) at (-1, 0.8) {};
                \node[circle, fill=purple!80!black, inner sep=1.8pt] (H1) at (0.4, -1.2) {};
                \node[circle, fill=purple!80!black, inner sep=1.8pt] (H2) at (0.4, -0.4) {};
                \node[circle, fill=purple!80!black, inner sep=1.8pt] (H3) at (0.4, 0.4) {};
                \node[circle, fill=purple!80!black, inner sep=1.8pt] (H4) at (0.4, 1.2) {};
                \node[circle, fill=red!70, inner sep=1.8pt] (O1) at (1.8, -0.5) {};
                \node[circle, fill=red!70, inner sep=1.8pt] (O2) at (1.8, 0.5) {};
                \foreach \i in {1,2,3} \foreach \j in {1,2,3,4} \draw[very thin, purple!50] (I\i) -- (H\j);
                \foreach \i in {1,2,3,4} \foreach \j in {1,2} \draw[very thin, purple!50] (H\i) -- (O\j);
            \end{scope}

            \node[font=\sffamily\scriptsize, align=center] at (0.7, -5.1) {\textbf{\small Agent Controller $\pi_{\theta, i}$}\\[0.8mm]Reward: $R_{\text{local}} + R_{\text{neighbor}} + R_{\text{global}}$};

            \node[rectangle, rounded corners=4pt, thick, dashed, draw=teal!80!black, fill=teal!4, minimum width=4.8cm, minimum height=1.35cm] (Ward) at (0, -7.35) {};
            \node[font=\sffamily\scriptsize\bfseries, text=teal!90!black] at (0, -6.85) {Ward Facility $i$};

            \draw[thick, fill=white, draw=black!70] (-2.1, -7.7) rectangle (-0.5, -7.15);
            \foreach \x in {-1.7, -1.3, -0.9} \draw[thick, draw=black!70] (\x, -7.7) -- (\x, -7.15);
            \node[font=\sffamily\scriptsize, text=black!85] at (-1.3, -7.9) {Queue $Q_i$};

            \draw[thick, fill=teal!20, draw=teal!80!black, rounded corners=1pt] (0.3, -7.7) rectangle (2.1, -7.15);
            \node[font=\sffamily\scriptsize\bfseries, text=teal!90!black, align=center] at (1.2, -7.425) {Beds: $c_i$ / $\rho_i(t)$};
            \draw[->, thick, draw=black!70] (-0.5, -7.425) -- (0.3, -7.425);

            \node[font=\sffamily\scriptsize, align=center] (InFlow) at (-6.9, -7.425) {\textbf{Patient Intake}\\[0.4mm]$\lambda_i$};

            \node[rectangle, rounded corners=2pt, thick, draw=orange!80!black, fill=orange!15, minimum width=2.3cm, minimum height=0.9cm, align=center, font=\sffamily\scriptsize] (Valve) at (-4.6, -7.425) {\textbf{Expert Action Mask}\\[0.4mm]Prior Path Filter};

            \draw[->, thick, orange!80!black] (InFlow.east) -- (Valve.west);
            \draw[->, thick, orange!80!black] (Valve.east) -- (-2.1, -7.425);

            \node[rectangle, rounded corners=2pt, thick, draw=green!60!black, fill=green!15, minimum width=2.4cm, minimum height=0.9cm, align=center, font=\sffamily\scriptsize] (Throttle) at (4.6, -7.425) {\textbf{Service Speedup}\\[0.4mm]$\mu_i' = (1 + \alpha_i) \mu_i$};

            \node[font=\sffamily\scriptsize, align=center] (OutFlow) at (6.9, -7.425) {\textbf{Discharge}\\[0.4mm]$\mu_i'$};
            \draw[->, thick, green!50!black] (2.1, -7.425) -- (Throttle.west);
            \draw[->, thick, green!50!black] (Throttle.east) -- (OutFlow.west);

            \draw[->, thick, draw=blue!70!black, line width=1.0pt] (0, -6.67) -- (0, -5.77) node[midway, right, font=\sffamily\scriptsize, text=blue!80!black, xshift=-0.5mm] {State $o_i(t)$};

            \node[font=\sffamily\scriptsize, align=center, text=blue!80!black] (FPIn) at (-5.6, -5.1) {\textbf{Neighbor Intents}\\[0.4mm]$\bigoplus \text{FP}_j(t)$};
            \draw[->, thick, draw=blue!70!black, line width=1.0pt] (FPIn) -- (-2.9, -5.1);

            \draw[->, thick, draw=orange!80!black, line width=1.1pt] (-1.5, -5.77) to[out=225, in=75] node[midway, above left, font=\sffamily\small\bfseries, text=orange!80!black, xshift=2mm, yshift=2mm] {$\lambda_i$-Policy} (Valve.north);

            \draw[->, thick, draw=green!50!black, line width=1.1pt] (1.5, -5.77) to[out=315, in=105] node[midway, above right, font=\sffamily\small\bfseries, text=green!50!black, xshift=-2mm, yshift=2mm] {$\alpha_i$-Policy} (Throttle.north);

            \draw[dashed, thick, draw=blue!70!black, ->] (MedSurg.south) -- (0.8, -4.0) node[pos=0.82, right, font=\sffamily\scriptsize\bfseries, text=blue!80!black, xshift=1mm] {Embedded Autonomous Agent Loop};

        \end{tikzpicture}
    }
    \caption{\textbf{Physics-Informed Multi-Agent Coordination Architecture.} \textbf{(a) Macro-Level:} The hospital BCMP queueing network calibrated on MIMIC-IV patient trajectories serves as the structural topology and resource capacity boundary. \textbf{(b) Micro-Level:} Closed-loop system architecture for each departmental agent controller at clinical node $i$. The decentralized neural actor monitors local ward telemetry $o_i(t)$ alongside neighbor action fingerprints ($\text{FP}_j$). To relieve operational congestion, the actor governs two control variables: adjusting service rate acceleration ($\alpha$-Policy) to expand bed discharge capacity, and modulating cooperative patient transfer routing ($\lambda$-Policy). Meanwhile, an expert prior action mask ensures topological safety by filtering medically unfeasible transitions from the exploration space.}
    \label{fig:bcmp_topology}
    \vspace{-1.5em}
\end{figure}

As depicted in Figure \ref{fig:bcmp_topology}, external patients enter the hospital according to class-specific arrival rates $\gamma_{i,k}$, and navigate between departments governed by an intrinsic transition probability matrix $P = [p_{(j,r),(i,k)}]$, where $p_{(j,r),(i,k)} \in [0, 1]$ denotes the transition probability of a class-$r$ patient transferring from node $j$ to node $i$ as class $k$. Under queueing mechanics \cite{baskett1975open}, traffic conservation dictates that the effective macroscopic arrival rate $\lambda_{i,k}$ at any department node $i$ satisfies the linear equilibrium:
\begin{equation}
    \label{eq:traffic}
    \lambda_{i,k} = \gamma_{i,k} + \sum_{j \in \mathcal{N}} \sum_{r \in \mathcal{K}} \lambda_{j,r} \cdot p_{(j,r),(i,k)}, \quad \forall i \in \mathcal{N}, \forall k \in \mathcal{K}.
\end{equation}
Because every hospital patient pathway eventually culminates in clinical discharge or transfer to an external absorbing state, the transition matrix across internal clinical nodes is substochastic with spectral radius $\rho(P) < 1$. This ensures that Eq.~\eqref{eq:traffic} admits a unique solution for $\lambda_{i,k}$, yielding the baseline offered workload intensity $a_i = \sum_{k \in \mathcal{K}} (\lambda_{i,k}/\mu_{i,k})$ for each department $i$, where $\mu_{i,k}$ denotes the class-specific baseline service rate. Correspondingly, the aggregate arrival rate is $\lambda_i = \sum_{k \in \mathcal{K}} \lambda_{i,k}$, and the aggregate departmental service rate $\mu_i$ satisfies $\mu_i = \lambda_i / a_i$.

\subsection{Physical Capacity Bounds and Stability Redlines}

To prevent severe queue accumulation and ensure that state transitions remain positive recurrent, Foster-Lyapunov stability theory dictates that each department's total offered traffic must not exceed its physical capacity limit $c_i$ (total beds or treatment stations). Thus, the operational traffic utilization $\rho_i$ must satisfy:
\begin{equation}
    \rho_i = \frac{a_i}{c_i} = \frac{1}{c_i} \sum_{k \in \mathcal{K}} \frac{\lambda_{i,k}}{\mu_{i,k}} < 1, \quad \forall i \in \mathcal{N}.
\end{equation}
To determine baseline structural bed capacities $c_i^*$ from historical empirical data, our system incorporates a Quality-and-Efficiency-Driven (QED) capacity calculation combining iterative Erlang-C delay probability verification with the Halfin-Whitt square-root staffing rule. Specifically, the initial capacity estimate is calculated as $c_i^{(0)} = \max(1, \lceil a_i + \beta_i \sqrt{a_i} \rceil)$. By enforcing targeted quality thresholds ($\epsilon_{\text{ED}}=0.01$, $\epsilon_{\text{ward}}=0.05$), the physics prior establishes realistic boundaries on hardware scaling, converging to the calibrated structural capacity $c_i^*$.

\subsection{The Breakdown of Stationary Product Forms}
While Eq.~\eqref{eq:traffic} defines theoretical demand, real hospital nodes possess finite queuing capacities. When a department experiences acute surges and bed occupancy approaches $c_i$, physical blocking occurs. In accordance with BCMP network conventions, departments operate under First-Come, First-Served (FCFS) service discipline; when concurrent transfers exceed available capacity $c_i$, incoming patients queue in order of their arrival timestamps, temporarily boarding in upstream units until downstream beds are vacated. Consequently, traditional stationary transition constants $p_{(i,k),(j,r)}$ abruptly degrade into complex, time-varying dynamic routing functions $p_{(i,k),(j,r)}(\mathbf{n})$ driven by the immediate global congestion vector $\mathbf{n} = (\mathbf{n}_1, \dots, \mathbf{n}_M)$.

This state-dependent routing disrupts local balance conditions, precluding the derivation of tractable analytical equilibrium formulas. In the presence of an expanding joint state space, stationary analytical models become insufficient for optimal dispatch control. Nevertheless, the structural topology of the network, specifically the directional routing graph, baseline service capacities $\mu_{i,k}$, and utilization limits $\rho_i < 1.0$, supplies an informative physical prior. We utilize this architectural domain knowledge to parameterize our decentralized multi-agent coordination system, linking static analytical bounds with active cooperative governance.

\vspace{-1em}
\section{Multi-Agent Coordination Framework}

To navigate the dynamic complexities of finite-capacity queueing networks, we formulate hospital patient flow management as an intelligent multi-agent collaboration problem.

\subsection{Why Multi-Agent Over Centralized RL?}
A primary structural consideration when modeling hospital congestion is why a decentralized multi-agent architecture is preferable to a monolithic centralized controller (such as a single global PPO or DQN policy). In operational institutions, centralized policy execution faces three practical constraints:
\begin{itemize}
    \item \textbf{Local Observations and Privacy:} Departmental supervisors observe real-time parameters locally within their respective wards (e.g., immediate queue backlogs and specialized bed utilization). Streaming raw clinical variables across all hospital units to a global processor creates unnecessary communication bandwidth costs and conflicts with administrative privacy boundaries.
    \item \textbf{Naturally Conflicting Objectives:} Departments operate with disparate operational goals. The ED seeks rapid admission evacuation to keep reception bay doors open; ICUs operate under rigorous admission thresholds to reserve mechanical ventilators for sudden surgeries; general wards strive to smooth nursing shifts and stabilize occupancy. A centralized scalar reward invariably marginalizes localized department priorities, triggering systemic instability.
    \item \textbf{Decentralized Actions Under Coupled Constraints:} Resource interventions, including discharge acceleration and admission control, are managed locally by departmental directors. However, these localized decisions intersect across topological pathways, as an ICU transfer rejection forces emergency department boarding and triggers upstream congestion cascades.
\end{itemize}
Thus, hospital operational dynamics represent decentralized decision-making under coupled constraints. Assigning autonomous agents to individual departmental nodes preserves administrative boundaries, conforms localized actions to domain limitations, and supports inter-agent routing negotiation.

\subsection{Dec-POMDP Formulation and Theoretical Rationale}
We mathematically formulate the hospital-wide coordination problem as a Decentralized Partially Observable Markov Decision Process (Dec-POMDP), defined by the tuple $\mathcal{M} = \langle \mathcal{N}, \mathcal{S}, \mathcal{A}, \mathcal{T}, \mathcal{R}, \Omega, \mathcal{O}, \gamma \rangle$, where $\mathcal{N}=\{1, \dots, M\}$ represents the departmental agent ensemble, $\mathcal{S}$ is the global environment state space, $\mathcal{A} = \prod_{i \in \mathcal{N}} \mathcal{A}_i$ and $\Omega = \prod_{i \in \mathcal{N}} \Omega_i$ denote the joint action and joint observation spaces across all departments, $\mathcal{T}: \mathcal{S} \times \mathcal{A} \times \mathcal{S} \to [0, 1]$ is the state transition probability function, $\mathcal{O}: \mathcal{S} \times \mathcal{A} \times \Omega \to [0, 1]$ is the observation emission probability function, $\mathcal{R} = \{R_i\}_{i \in \mathcal{N}}$ represents the set of departmental reward functions, and $\gamma \in [0, 1)$ denotes the temporal discount factor.

\subsubsection{Local Observation Space $\Omega_i$ and Rationale.}
Due to partial observability, agent $i \in \mathcal{N}$ does not access the global state $\mathbf{n} \in \mathcal{S}$. Instead, at decision step $t$, agent $i$ receives a localized observation vector $o_i(t) \in \Omega_i$. Rather than streaming high-dimensional telemetry across nodes, local clinical indicators are encoded into a compact 5-dimensional representation:
\begin{equation}
    o_i(t) = \left[ o_{i}^{\text{occ}}(t), \, o_{i}^{\text{queue}}(t), \, o_{i}^{\text{avail}}(t), \, o_{i}^{\text{sla}}(t), \, o_{i}^{\text{alpha}}(t) \right],
\end{equation}
which capture the bed occupancy ratio, normalized waiting queue depth, available capacity proportion, Service Level Agreement (SLA) violation warning flags (triggered when patient waiting delay exceeds standard thresholds), and the currently active discharge acceleration intensity, respectively. From a queueing theory perspective, these five indicators perfectly encapsulate the instantaneous traffic utilization parameter $\rho_i(t)$ and queue derivative $dL_{q,i}/dt$, providing sufficient state representation for decentralized action inference without requiring full visibility of distant hospital wards.

\subsubsection{Discrete Cooperative Action Space $\mathcal{A}_i$ and Rationale.}
Each departmental agent $i$ exercises dual-channel intervention authority over local patient flow through a discrete action space $\mathcal{A}_i = \mathcal{A}_i^{\alpha} \times \mathcal{A}_i^{\lambda}$:
\begin{itemize}
    \item \textbf{Accelerated Discharge Control ($\alpha$-Policy, 4 Levels):} The agent selects a service acceleration factor $\alpha_i \in \{0, 0.05, 0.10, 0.15\}$ to proactively mobilize auxiliary clinical staff, accelerating local treatment completion from baseline $\mu_i$ to $\mu_i' = (1 + \alpha_i)\mu_i$. The $15\%$ ceiling conforms to realistic clinical staffing flexibility, while resource expenditures and safety risks are strictly regulated via management and premature-discharge penalties ($W_{4,i}$ and $W_{3,i}$ in Eq.~\eqref{eq:local_reward}).
    \item \textbf{Cooperative Routing Integration ($\lambda$-Policy, 5 Levels):} The agent selects an admission gating factor $\lambda_i \in \{0.0, 0.25, 0.50, 0.75, 1.0\}$ across five discrete levels to scale incoming external and upstream transfer demand ($\lambda_i \cdot \Lambda_i$). Here, $\lambda_i=1.0$ maintains unconstrained intake, $\lambda_i=0.0$ executes emergency throttling during acute congestion, and intermediate tiers ($\{0.25, 0.50, 0.75\}$) provide graduated multi-stage regulation (admitting 25\%, 50\%, or 75\%) that avoids binary switching oscillations. Conditioned on internal queue/bed saturation ($o_i^{\text{occ}}, o_i^{\text{queue}}$) and communicated upstream neighbor pressure ($\max_{j \in \mathcal{N}_i^{\text{up}}} q_j$), agents down-modulate $\lambda_i$ to decompress saturated wards and coordinate diversion to step-down units.
\end{itemize}
\textit{Theoretical Design Rationale:} Discretizing action dimensions into bounded levels prevents non-stationary policy gradient oscillations during multi-agent negotiation, stabilizing spatial coordination while providing interpretable operational decisions for clinical management.

\subsection{Topology-Aware Communication and Cooperative Negotiation}
In standard independent MARL, agents updating policies locally using only internal observations $o_i(t)$ experience environmental non-stationarity, because as one unit alters its routing policy, neighboring arrival rates $\lambda_{i}(t)$ shift unpredictably. We address this communication discrepancy by implementing a topology-aware coordination mechanism grounded in MA2C (Mean-field Advantage Actor-Critic) principles under a Centralized Training with Decentralized Execution (CTDE) architecture \cite{chu2019multi}.

\subsubsection{Action Fingerprint Communication Strategy.}
Instead of establishing unconstrained all-to-all communication networks that scale quadratically with hospital departments, agents share information along directed topological edges derived from our physics prior (Figure \ref{fig:bcmp_topology}). During decentralized execution, agent $i$ actively communicates with its adjacent topological neighbors by broadcasting its action fingerprint ($\text{FP}_i$).

Specifically, each agent globally monitors an Exponential Moving Average (EMA) of its historical policy output distributions, updating via $\text{FP}_i \leftarrow (1 - \alpha_{\text{fp}}) \text{FP}_i + \alpha_{\text{fp}} \pi_i(a_i|o_i)$, with smoothing factor $\alpha_{\text{fp}} = 0.1$. At step $t$, agent $i$ concatenates the incoming action fingerprints of its direct topological neighbors $j \in \mathcal{N}_i$ into its operational observation space:
\begin{equation}
    \widetilde{o}_i(t) = \big[ o_i(t); \, \bigoplus\nolimits_{j \in \mathcal{N}_i} \text{FP}_j(t) \big].
\end{equation}
This communicative coupling helps mitigate multi-agent partial observability and stabilizes collaborative training by furnishing adjacent nodes with real-time feedback regarding neighborhood queue trajectories and policy shifts.

\subsubsection{Emergence of Cooperative Routing Negotiation.}
By combining topological action fingerprints with decentralized policy actors, patient routing ceases to be a top-down operational directive and emerges natively from dynamic multi-agent negotiation. For instance, when an emergency influx hits the ED agent, its rising queue backlog shifts its action fingerprint $\text{FP}_{\text{ED}}$, broadcasting acute upstream pressure to neighboring units. In response, downstream ICU agents sensing the impending arrival surge via communicated fingerprints can cooperatively restrict non-critical elective admissions ("negotiate restriction"), while step-down general medicine wards initiate discharge acceleration ($\alpha_j > 0$) to preemptively clear beds for step-down transfers ("negotiate accommodation"). Thus, hospital routing trajectories dynamically stabilize through inter-agent cooperative problem-solving rather than rigid predefined matrices.

\subsection{Expert-Guided Action Masking and Topological Safety}
While cooperative MARL effectively mitigates operational bottlenecks, unconstrained actors maximizing queue rewards could attempt medically unfeasible transfers (e.g., routing high-acuity patients to general wards lacking resuscitation infrastructure). To enforce clinical validity, we establish an online expert action masking mechanism derived from MIMIC-IV attending protocols.

During policy execution for each agent $i \in \mathcal{N}$, candidate actions $a_i \in \mathcal{A}_i$ given observation $o_i \in \Omega_i$ are validated against the empirical clinical transition manifold. Any pathway violating critical care triage guidelines \cite{blanch2016triage} or lacking clinical precedence is explicitly pruned before action sampling:
\begin{equation}
    \pi_{\text{safe}}(a_i | o_i) = \frac{\pi_{\theta, i}(a_i | o_i) \cdot \mathbb{I}_{\text{valid}}(a_i)}{\sum_{a' \in \mathcal{A}_i} \pi_{\theta, i}(a' | o_i) \cdot \mathbb{I}_{\text{valid}}(a')},
\end{equation}
where $\theta$ denotes the trainable weight parameters of the policy neural network $\pi_{\theta, i}$, and $\mathbb{I}_{\text{valid}}(a_i) \in \{0, 1\}$ indicates clinically permissible transitions for candidate action $a_i \in \mathcal{A}_i$. Filtering invalid pathways confines exploratory routing to safe clinical envelopes, ensuring adherence to documented care standards.

\vspace{-1em}
\section{Reward Decomposition and Multi-Agent Credit Assignment}

In complex multi-agent environments, utilizing a unified global scalar reward across all actors complicates multi-agent credit assignment, as individual agents struggle to discern how local actions impact hospital-wide performance. Conversely, relying strictly on localized queue incentives encourages uncoordinated behaviors, such as diverting bottlenecked queues to adjacent wards. We reconcile localized efficiency with system-wide stability by formulating a spatially hierarchical reward decomposition architecture.

At any step $t$, the collaborative feedback signal delivered to agent $i$ is separated into three structural layers:
\begin{equation}
    R_{i}(t) = \omega_{\text{local}} \, R_{\text{local}, i}(t) + \omega_{\text{neighbor}} \, R_{\text{neighbor}, i}(t) + \omega_{\text{global}} \, R_{\text{global}}(t).
\end{equation}
This hierarchical structure decomposes credit assignment across spatial scales: the local term maintains throughput under resource costs, the neighborhood term suppresses cross-ward spillover, and the global derivative guides macroscopic convergence.

\subsection{Local Domain Objective}
The local objective aligns agent behavior with intra-departmental physical efficiency, penalizing immediate queue accumulation while regulating excessive management intervention:
\begin{equation}
    \label{eq:local_reward}
    R_{\text{local}, i}(t) = - \left( W_{1, i}(t) + W_{2, i}(t) + W_{3, i}(t) + W_{4, i}(t) - W_{6, i}(t) \right).
\end{equation}
Here, the individual operational penalties are formulated as follows: \emph{queue backlog penalty} $W_{1, i}(t) = 1.0 \times Q_i(t)$, which directly penalizes waiting queue depth $Q_i(t)$ rather than normal admitted beds; \emph{utilization redline penalty} $W_{2, i}(t) = 1.0 \times \max(0, u_i(t) - 0.85)^2$, which applies a quadratic penalty when bed utilization $u_i(t)$ breaches the critical $85\%$ capacity redline; \emph{premature-discharge risk} $W_{3, i}(t) = 1.5 \times (\exp(2.0 \alpha_i(t)) - 1)$, an exponential barrier against unsafe discharge acceleration; \emph{management cost} $W_{4, i}(t) = 0.2 \times (c_i \alpha_i(t))$, accounting for operational staffing expenses; and \emph{local credit bonus} $W_{6, i}(t) = 0.3 \times \mathbb{I}(\alpha_i(t) > 0) \max(0, Q_i(t-1) - Q_i(t))$, providing direct positive feedback when active acceleration successfully reduces local waiting lists.

\subsection{Neighbor Collaborative Objective}
To discourage localized optimization strategies that simply relieve internal ward congestion by displacing bottlenecked queues onto directly connected departments, we institute a spatial collaborative reward layer regulated by a neighborhood spatial discount factor ($\gamma_{\text{coop}} = 0.9$):
\begin{equation}
    R_{\text{neighbor}, i}(t) = \frac{\gamma_{\text{coop}}}{|\mathcal{N}_i|} \sum_{j \in \mathcal{N}_i} R_{\text{local}, j}(t).
\end{equation}
By incorporating the averaged immediate operational feedback of topologically connected peers, this cooperative objective incentivizes departmental agents to maintain receptive patient transition pathways and coordinate admission timing without requiring global communication bandwidth.

\subsection{Global Systemic Objective}
Finally, to anchor all agents to the supreme clinical imperative of hospital-wide flow fluidity, we incorporate a globally shared incremental delay objective based on total network patient delay $D(t)$:
\begin{equation}
    R_{\text{global}}(t) = - 0.2 \times \left( D(t) - D(t-1) \right).
\end{equation}
By evaluating the first-order derivative of system delay rather than absolute cumulative delay, this term broadcasts continuous, zero-mean credit assignment signals across the multi-agent ensemble, positively reinforcing cooperative strategies that accomplish macroscopic delay contraction.

\vspace{-1em}
\section{Experiments and Empirical Evaluation}

\subsection{Experimental Configuration and EHR Calibration}
Our multi-agent simulation environments are engineered within a high-fidelity Discrete Event Simulation (DES) engine calibrated on the complete MIMIC-IV inpatient cohort of approximately $60\,000$ patient admissions \cite{johnson2023mimic}, from which empirical arrival intensities, service distributions, and transition matrices $P$ are parameterized. To evaluate collaborative resilience under operational stress, all experiments simulate stochastic arrival fluctuations under an acute external demand surge multiplier of 1.25.

During the training regimen, the decentralized MA2C agents optimize over 250 simulation episodes, where each episodic epoch represents a continuous operational duration of 4000 hours of simulated hospital workflow. To guarantee broad multi-agent exploration across quantized negotiation tables and thwart premature policy freeze, the policy entropy coefficient is established at 0.05.

\subsection{Comparative Analysis against Multi-Agent Baselines}
To evaluate whether performance improvements stem from genuine multi-agent coordination rather than mere mathematical capacity expansion, we contrast our proposed architecture against four rigorous multi-agent and analytical benchmarks: (1) Static Markovian Baseline ($\alpha = 0$), which models stationary unassisted M/M/c evolution under BCMP assumptions; (2) Greedy Heuristic Dispatcher, which grants individual departmental controllers an identical $15\%$ acceleration budget ($\alpha \le 0.15$) operated via decentralized myopic threshold rules without inter-agent communication; (3) Independent MARL Baseline (IA2C), where independent Advantage Actor-Critic agents optimize individualized policies solely from local observations $o_i(t)$ without topology-aware action fingerprints or neighbor reward coupling; and (4) Cooperative MA2C (our proposed framework). Both multi-agent approaches (IA2C and MA2C) are evaluated under deterministic (greedy) and exploratory (stochastic) execution modes to assess policy stability under different action selection regimes.

\begin{table}[htbp]
    \centering
    \caption{System Performance and Departmental Bottleneck Queues across MAS Strategies}
    \label{tab:evaluation_results_with_bottlenecks}
    \renewcommand{\arraystretch}{1.1}
    \resizebox{\textwidth}{!}{
        \begin{tabular}{lcccccc}
            \toprule
            \multirow{2}{*}{\textbf{Policy Strategy}} & \multirow{2}{*}{\textbf{Joint Reward}} & \multirow{2}{*}{\textbf{Cumulative Total Delay (hrs)}} & \multirow{2}{*}{\textbf{Mean $\alpha$}} & \multicolumn{3}{c}{\textbf{Mean Bottleneck Queue Depth ($L_q$)}}                                               \\
            \cmidrule(lr){5-7}
                                                      &                                        &                                                        &                                         & \textbf{Medicine Ward}                                           & \textbf{Med/Surg} & \textbf{Emergency (ED)} \\
            \midrule
            Cooperative MA2C (Greedy)                 & $-0.278 \pm 0.000$                     & $3974.0 \pm 481.0$                                     & $0.044$                                 & $0.043$                                                          & $0.009$           & $0.636$                 \\
            Cooperative MA2C (Stochastic)             & $-0.290 \pm 0.000$                     & $\mathbf{2081.2 \pm 571.7}$                            & $0.075$                                 & $\mathbf{0.000}$                                                 & $\mathbf{0.005}$  & $\mathbf{0.295}$        \\
            Independent A2C (Greedy)                  & $-0.274 \pm 0.072$                     & $463038.7 \pm 243695.6$                                & $0.061$                                 & $2.996$                                                          & $2.954$           & $1.684$                 \\
            Independent A2C (Stochastic)              & $-0.608 \pm 0.001$                     & $2789.7 \pm 642.8$                                     & $0.075$                                 & $\mathbf{0.000}$                                                 & $\mathbf{0.000}$  & $0.630$                 \\
            Greedy Heuristic ($\alpha \le 0.15$)      & ---                                    & $43148.5$                                              & $0.004$                                 & $2.242$                                                          & $1.869$           & $0.850$                 \\
            Static Markovian Baseline ($\alpha = 0$)  & ---                                    & $117478.1$                                             & ---                                     & $12.954$                                                         & $1.869$           & $0.850$                 \\
            \bottomrule
        \end{tabular}
    }
    \vspace{-1em}
\end{table}

As detailed in Table \ref{tab:evaluation_results_with_bottlenecks}, unguided hospital operations under the Static Markovian Baseline yield a severe cumulative delay of $117\,478.1$ hours over the 4000-hour horizon, driven primarily by acute persistent congestion within the primary Medicine Ward ($L_q = 12.954$). Implementing the Greedy Heuristic Dispatcher moderately reduces cumulative delays to $43\,148.5$ hours, yet fails to resolve systemic bottlenecks due to myopic reactive interventions ($\text{Mean }\alpha = 0.004$) that lack spatial coordination with upstream patient traffic surges.

Evaluating the Independent MARL baseline (IA2C) under deterministic versus exploratory execution reveals acute policy sensitivity to action selection regimes. Under stochastic execution, independent actors achieve meaningful delay reduction ($2\,789.7 \pm 642.8$ hours) and successfully clear queues across General Medicine ($L_q = 0.000$) and Med/Surg ($L_q = 0.000$) units by actively intervening ($\text{Mean }\alpha = 0.075$). However, when operating deterministically under greedy execution without exploratory noise, IA2C experiences catastrophic operational collapse: cumulative delays diverge to $463\,038.7 \pm 243\,695.6$ hours accompanied by severe bottleneck accumulation across all wards ($L_q = 2.996$ in Medicine and $L_q = 2.954$ in Med/Surg). This instability directly illustrates our theoretical assertions: independent agents learning without neighborhood communicative fingerprints encounter severe environmental non-stationarity, trapped in localized optimization deadlocks and oscillatory routing conflicts when exploratory randomness is removed.

In contrast, the Cooperative MA2C architecture demonstrates superior and robust congestion mitigation across both evaluation modes, avoiding the instability of uncoordinated learning. Under exploratory stochastic evaluations, our multi-agent framework confines cumulative hospital delay to $\mathbf{2081.2 \pm 571.7}$ hours ($3974.0 \pm 481.0$ hours under greedy execution), yielding an approximate 56-fold reduction compared to static Markovian approximations and outperforming independent MARL actors across both operational regimes. This performance is achieved with consistent overall discharge acceleration ($\text{Mean }\alpha = 0.075$). By leveraging inter-agent communicative coordination and topological physical priors, cooperative policies actively divert secondary transfers away from saturated nodes, eliminating queue accumulation across the General Medicine ward ($L_q = \mathbf{0.000}$) and nearly eliminating it in Med/Surg units ($L_q = \mathbf{0.005}$), while registering the lowest bottleneck queue depth within the Emergency Department ($L_q = \mathbf{0.295}$).

\subsection{Structural Ablation Studies}
We quantify the contribution of each architectural component within our proposed framework through structural ablation experiments by isolating core mechanisms from the complete Cooperative MA2C model. To isolate architectural component efficacy from action-selection variance across execution regimes, all ablation variants and the reference model are evaluated across an independent set of standardized random evaluation seeds (accounting for the empirical variance between Table \ref{tab:evaluation_results_with_bottlenecks} and Table \ref{tab:ablation_results}).

\begin{table}[htbp]
    \centering
    \caption{Structural Ablation Evaluation of Multi-Agent Framework Components}
    \label{tab:ablation_results}
    \renewcommand{\arraystretch}{1.1}
    \resizebox{\textwidth}{!}{
        \begin{tabular}{lccc}
            \toprule
            \textbf{Ablated Configuration Architecture}      & \textbf{Cumulative Delay (hrs)} & \textbf{Mean Speedup ($\alpha$)} & \textbf{ICU Premature Discharge Rate (\%)} \\
            \midrule
            \textbf{Full Cooperative MA2C (Proposed)}        & $\mathbf{2835.2 \pm 972.1}$     & $0.044$                          & $0.211$                                    \\
            \midrule
            (a) w/o Neighbor Action Fingerprints             & $3024.5 \pm 893.0$              & $0.037$                          & $0.227$                                    \\
            (b) w/o Cooperative Routing ($\lambda$-disabled) & $226231.1 \pm 23861.5$          & $0.051$                          & $0.368$                                    \\
            (c) w/o Dynamic Acceleration ($\alpha$-disabled) & $3990.8 \pm 877.6$              & $0.000$                          & $0.212$                                    \\
            (d) w/o Expert Prior Action Masking              & $3108.9 \pm 676.8$              & $0.044$                          & $0.190$                                    \\
            \bottomrule
        \end{tabular}
    }
    \vspace{-1em}
\end{table}

The ablation findings presented in Table \ref{tab:ablation_results} substantiate our architectural design choices: removing neighbor action fingerprints (a) elevates cumulative delay from $2\,835.2$ to $3\,024.5$ hours while slightly dampening discharge acceleration ($\text{Mean }\alpha = 0.037$), indicating that active topology-aware signaling assists agents in anticipating incoming traffic shifts and mitigating Dec-POMDP non-stationarity; disabling cooperative routing (b) drives overall delay into acute divergence ($226\,231.1$ hours) alongside an elevated premature discharge rate ($0.368\%$), demonstrating that localized capacity scaling alone cannot bypass structural network bottlenecks without spatial routing negotiation; disabling dynamic acceleration (c) increases cumulative delay to $3\,990.8$ hours ($\text{Mean }\alpha = 0.000$), confirming that auxiliary capacity elasticity is essential to relieve peak localized surges and maintain stable throughput; finally, removing expert prior action masking (d) increases system delay to $3\,108.9$ hours with an observed premature discharge rate of $0.190\%$. Without explicit empirical priors bounding the transition manifold, unguided actors adopt misaligned routing strategies, such as overly restrictive transfer policies that artificially compress local ICU throughput while displacing congestion upstream, confirming the necessity of expert action masking to ensure balanced operational efficiency and adherence to documented clinical standards.

\subsection{Clinical Pathway Safety and Anomaly Verification}
A resilient medical AI system must enhance operational scheduling without introducing anomalous clinical pathways. To verify clinical safety, we monitored all patient trajectories throughout simulation execution, quantifying two major pathway violations against historical empirical baselines in MIMIC-IV: ping-pong transfers, representing cyclical, redundant ward transitions (e.g., Ward A $\to$ Ward B $\to$ Ward A) that sever care continuity; and premature ICU discharges, capturing high-risk events where unstable critical care patients are discharged directly out of Intensive Care Units without traversing required step-down wards.

\begin{table}[htbp]
    \centering
    \caption{Empirical Validation of Clinical Pathway Safety against Real Hospital Records}
    \label{tab:clinical_pathway_safety}
    \renewcommand{\arraystretch}{1.1}
    \resizebox{\textwidth}{!}{
        \begin{tabular}{lcc}
            \toprule
            \textbf{Operational Policy Benchmark}                  & \textbf{Ping-Pong Transfer Rate (\%)} & \textbf{Premature ICU Discharge Rate (\%)} \\
            \midrule
            Real Historical Data (MIMIC-IV Records)                & 7.375                                 & 2.585                                      \\
            \midrule
            Independent MARL Baseline (IA2C, Greedy Execution)     & 2.769                                 & 0.298                                      \\
            Independent MARL Baseline (IA2C, Stochastic Execution) & 0.674                                 & 0.095                                      \\
            Cooperative MA2C (Greedy Execution)                    & 1.727                                 & 0.214                                      \\
            Cooperative MA2C (Stochastic Execution)                & 1.736                                 & 0.211                                      \\
            \bottomrule
        \end{tabular}
    }
    \vspace{-1em}
\end{table}

As documented in Table \ref{tab:clinical_pathway_safety}, operational trajectories in MIMIC-IV historical records exhibit a baseline ping-pong transition rate of $7.375\%$ and a premature ICU discharge incidence of $2.585\%$. Here, a ping-pong transfer is identified when a patient revisits the same ward $\ge 2$ times within a single admission ($A \to B \to A$), while premature ICU discharge denotes transfers occurring before completing the department's calibrated Average Length of Stay ($\text{ALOS}_{\text{ICU}}$). All learned policies substantially reduce both anomaly types relative to historical baselines. Independent MARL actors (IA2C) achieve the lowest ping-pong transfer rates ($0.674\%$ under stochastic execution) due to conservative routing. Our Cooperative MA2C framework records moderately higher ping-pong rates ($1.727\%$ and $1.736\%$ under greedy and stochastic execution, respectively), reflecting its active cooperative routing strategy that redistributes patients across wards to relieve congestion bottlenecks. Importantly, both MA2C execution modes maintain premature ICU discharge rates ($0.214\%$ and $0.211\%$) comparable to IA2C stochastic ($0.095\%$) and substantially below historical baselines ($2.585\%$), confirming that the expert action mask effectively preserves clinical safety even under dynamic cooperative routing.

\vspace{-1em}
\section{Discussion \& Conclusion}

A common abstraction in healthcare operations modeling is the implicit assumption of a linear relationship between incoming patient traffic demand and emergent hospital congestion, namely, that a $10\%$ increase in patient arrivals will induce an approximately proportional $10\%$ expansion in waiting times.

However, queueing theory reveals a non-linear relationship. Canonical queuing formulations, such as the Erlang-C model for M/M/c multi-server queues and the Pollaczek--Khinchin formula for M/G/1 systems, indicate that expected waiting queue depths grow according to an asymptotic inverse relationship with departmental resource utilization:
\begin{equation}
    L_q \propto \frac{1}{1-\rho_i}.
\end{equation}

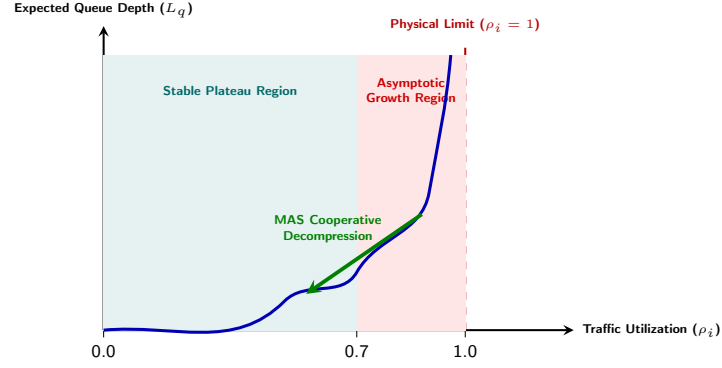
\begin{figure}[!h]
    \centering
    \vspace{-0.5em}
    \resizebox{0.58\textwidth}{!}{
        \begin{tikzpicture}[
                font=\sffamily\scriptsize,
                >=stealth
            ]
            \draw[->, thick] (0,0) -- (6.5,0) node[right, font=\sffamily\tiny\bfseries] {Traffic Utilization ($\rho_i$)};
            \draw[->, thick] (0,0) -- (0,4.2) node[above, font=\sffamily\tiny\bfseries] {Expected Queue Depth ($L_q$)};

            \draw (0,0) -- (0,-0.1) node[below] {0.0};
            \draw (3.5,0) -- (3.5,-0.1) node[below] {0.7};
            \draw (5.0,0) -- (5.0,-0.1) node[below] {1.0};

            \draw[red!70!black, dashed, thick] (5.0,0) -- (5.0,4.0) node[above, font=\sffamily\tiny\bfseries] {Physical Limit ($\rho_i = 1$)};

            \fill[teal!10] (0,0) rectangle (3.5,3.8);
            \fill[red!10] (3.5,0) rectangle (5.0,3.8);

            \node[text=teal!80!black, font=\sffamily\tiny\bfseries] at (1.75, 3.3) {Stable Plateau Region};
            \node[text=red!80!black, font=\sffamily\tiny\bfseries, align=center] at (4.25, 3.3) {Asymptotic\\Growth Region};

            \draw[blue!70!black, very thick] (0,0) to[out=5, in=225] (2.5,0.4) to[out=45, in=240] (3.5,0.8) to[out=60, in=260] (4.5,1.9) to[out=80, in=265] (4.8,3.8);

            \draw[->, thick, green!50!black, line width=1.5pt] (4.4,1.6) -- (2.8,0.5);
            \node[text=green!50!black, font=\sffamily\tiny\bfseries, align=center] at (3.1, 1.4) {MAS Cooperative\\Decompression};
        \end{tikzpicture}
    }
    \caption{Asymptotic queue escalation as department utilization $\rho_i \to 1^-$, illustrating how localized multi-agent interventions actively transition operations back into the stable operating plateau.}
    \label{fig:nonlinear_queue}
    \vspace{-1.5em}
\end{figure}

As illustrated in Figure \ref{fig:nonlinear_queue}, when a clinical unit operates within moderate utilization boundaries ($\rho_i \leq 0.7$), the department resides on a stable operating plateau where stochastic arrival increases yield negligible wait-time escalation. However, once operational utilization approaches the physical capacity limit ($\rho_i \to 1^-$), the denominator $(1-\rho_i)$ approaches zero, driving queue accumulation and waiting delay into a superlinear asymptotic surge. Within this congestive regime, minor arrival perturbations can trigger inter-departmental blocking cascades.

This relationship explains why stationary approximations degrade under heavy load and illustrates the role of multi-agent coordination in dynamic healthcare optimization. Under static modeling assumptions, high-demand wards such as General Medicine operate continuously near saturation ($\rho_i \approx 0.999$), generating an accumulated simulation delay of $117\,478.1$ hours. Because individual wards experience localized queue surges independently, neither centralized scheduling policies nor uncoordinated routing rules react effectively to these non-linear transitions without inducing inter-departmental bottlenecks.

Our approach mitigates these bottlenecks through cooperative multi-agent decision-making. By coupling localized departmental actors with topological action-fingerprint communication, the policy network maintains responsiveness to regional congestion gradients. Instead of requiring global staffing additions, agents negotiate localized interventions by combining slight service acceleration ($\text{Mean }\alpha = 0.075$) with cooperative patient routing diversion. This multi-agent policy action addresses the mathematics of the asymptotic curve: introducing temporary service speedup to a heavily utilized ward reduces its effective utilization below the non-linear inflection threshold ($\rho_i \to 1^-$), restoring operations to the linear delay regime. As a result, timed cooperative interventions accomplish a 56-fold reduction in simulation delay ($2\,081.2$ hours), validating the practical utility of decentralized multi-agent coordination.

\subsubsection{Limitations and Future Research Horizons.}
Although our proposed framework demonstrates effective physics-informed multi-agent coordination, open challenges remain. Currently, baseline service capacities $\mu_{i,t}$ rely on historical averages from MIMIC-IV and aggregated clinical stage identifiers, overlooking how real-time clinician workload, nursing fatigue, and workplace stress dynamically alter service efficiency. To bridge this simulation-to-reality gap, future research will explore a hierarchical hybrid framework coupling macroscopic MARL routing with microscopic behavioral simulation \cite{hadfi2016multilayered,hadfi2016holonic}. Specifically, embedding LLM-based cognitive agents (LLM-Agents) can simulate real-time clinician reasoning and staff behavioral dynamics under stress, offering a promising path toward interpretable decision support and resilient operational governance across complex healthcare institutions.

\begin{credits}
\subsubsection{\ackname} This work was supported by JST SPRING Grant Number\linebreak N01261400041 and JSPS KAKENHI Grant Number JP22H00533.

\subsubsection{Disclosure of Interests.} The authors have no competing interests to declare that are relevant to the content of this article.

\subsubsection{AI Declaration.} During the preparation of this work, the authors used ChatGPT to assist with English readability and phrasing. The authors reviewed and edited the content and take full responsibility for the publication.
\end{credits}

\bibliographystyle{splncs04}
\bibliography{main}
\end{document}